\documentclass[10pt,twocolumn,letterpaper]{article}
\usepackage{algorithm}
\usepackage{algpseudocode}
\usepackage{tikz}
\usetikzlibrary{arrows.meta,positioning,fit,calc}
\usepackage{xcolor}

\usepackage{cvpr}              
\definecolor{cvprblue}{rgb}{0.21,0.49,0.74}
\usepackage[pagebackref,breaklinks,colorlinks,allcolors=cvprblue]{hyperref}

\title{Towards Continuous Intelligence Growth: Self-Training, Continual Learning, and Dual-Scale Memory in SuperIntelliAgent}

\author{Jianzhe Lin\thanks{corresponding author:peter.lin@vicino.ai}, Zeyu Pan, Yun Zhu\thanks{independent researcher}, Ruiqi Song, Jining Yang\\
vicino.ai, https://vicino.ai/\\
code: https://github.com/jianzhelin/Superintelliagent \\
}

\begin{document}
\maketitle
\begin{abstract}
We introduce SuperIntelliAgent, an agentic learning framework that couples a trainable small diffusion model (the learner) with a frozen large language model (the verifier), enabling continual intelligence growth through self-supervised interaction. Unlike conventional supervised fine-tuning with annotated data, SuperIntelliAgent learns autonomously in an annotation-free manner: the learner generates candidate outputs, the verifier evaluates them via step-by-step reasoning, and the learner-verifier interaction loop produces chosen/rejected pairs for Direct Preference Optimization (DPO), transforming every input into a pseudo-training signal for continual self-improvement. The framework integrates a dual-scale memory mechanism—short-term, in-context memory that preserves reasoning traces across iterative refinement cycles, and long-term memory that consolidates acquired knowledge into model parameters through on-the-fly fine-tuning. To enhance optimization, a replay buffer selectively retains samples showing verifiable progress from failed to satisfied conditions and replays them as auxiliary supervision, reinforcing recent learning while bootstrapping adaptive curricula that accelerate intelligence acquisition. Designed to be infrastructure-agnostic, SuperIntelliAgent can be seamlessly integrated into existing agentic frameworks (e.g., autogen, semantic kernel, etc.), while simultaneously transforming ordinary inference cycles into lifelong optimization. We posit that agentic pairing constitutes the minimal reliable unit of growing intelligence, as paired feedback, augmented with partial-history replay, yields richer learning curricula, tighter preference alignment, and stronger generalization. With extremely few DPO pairs generated automatically by SuperIntelliAgent and used for lightweight fine-tuning, the learner performance improves across all benchmarks. This demonstrates that our proposed mechanism for continual intelligence accumulation offers a promising direction for future research and real-world model deployment.
\end{abstract}    
\section{Introduction}
\label{sec:intro}
Large-scale foundation models have achieved remarkable success, but remain fundamentally limited by data scarcity and static one-off training paradigms \cite{bommasani2021foundation, hoffmann2022chinchilla}. Most vision and multimodal systems are trained once on a massive, fixed dataset and then “frozen” for deployment, unable to continuously adapt or self-correct when faced with new environments \cite{alayrac2022flamingo, radford2021clip, ramesh2022dalle2}. This rigid training approach contrasts sharply with human intelligence, which learns continuously through feedback, reflection, and social interaction. Moreover, acquiring diverse, high-quality labeled data for every new domain is prohibitively expensive – especially for generative tasks like text-to-image reasoning \cite{abouelenin2025phi4multimodal, taori2020imagenetrealtough}. Without ongoing adaptation, deployed models inevitably drift away from real-world data distributions, causing performance degradation and brittle generalization over time \cite{wu2024continual, yu2024multimodalcl}. These challenges highlight the need for a new paradigm of continual self-improving learning that can convert everyday interactions, errors, and partial successes into fresh training signals automatically, without relying on external human supervision.

To eliminate dependence on human feedback, we propose an automatic preference-synthesis mechanism in which a frozen large language model (LLM) serves as a reasoning-based verifier for a trainable diffusion model. Inspired by recent self-critique approaches, the verifier uses chain-of-thought prompting to decompose each input prompt into semantically grounded sub-goals \cite{tian2025unigen}. For example, the prompt “generate a green banana” can be broken into verifiable conditions like object: banana and attribute: green. The diffusion learner’s output is then assessed against each condition via cross-modal entailment, producing a structured judgment vector that captures the fine-grained alignment between the intended attributes and the generated image. Over a sequence of self-reflection steps, the system generates “No" to "Yes” trajectories—initial samples that fail certain conditions progressively reflects and refine until all conditions are satisfied \cite{madaan2023selfrefine, shinn2023reflexion}. These trajectory pairs (before vs. after refinement) are treated as negative/positive examples for Direct Preference Optimization (DPO) \cite{rafailov2023dpo, hong2024orpo}. In this way, each generation–verification cycle yields a self-contained training signal, allowing the agent to construct its own preference dataset in real-time and improve itself without any human annotations.

We formalize the learner–verifier pair as an autonomous agent whose reasoning unfolds along a thread — a temporally coherent trajectory of generation, critique, and revision. Within each thread, the diffusion learner retains a short-term memory of intermediate hypotheses and the verifier’s feedback, iteratively refining its output until all chain-of-thought conditions are met. Across different threads, these transient “experience traces” are stored in a replay buffer, forming an ever-expanding curriculum of partial successes and corrections\cite{lin1992experiencereplay, rebuffi2017icarl, parisi2019continual, bengio2009curriculum, graves2017automatedcurriculum, florensa2017reversecurriculum, kaplan2023explosive}. The learner then performs asynchronous DPO training by sampling from this buffer, gradually distilling the transient reasoning patterns into long-term model parameters. This Auto-DPO process effectively interleaves inference-time reflection with training-time adaptation: while one batch of samples is being generated and verified, another batch (from earlier threads) is simultaneously used to fine-tune the model. Such a coupled design — short-term reasoning within threads and long-term consolidation across threads — is reminiscent of experience replay in reinforcement learning and enables continual, self-driven improvement without human intervention.

The proposed system is designed for seamless integration into existing agent frameworks (e.g. Microsoft’s Semantic Kernel and AutoGen) without requiring changes to their orchestration or messaging interfaces \cite{yao2022react, schick2023toolformer, wu2023autogen}. The learner–verifier pair functions as a self-contained, continuously learning component: the learner model improves autonomously using feedback signals synthesized entirely by the advisor (verifier). In deployment, this learning loop operates transparently alongside normal inference: every generation is automatically analyzed, critiqued, and adjusted based on internal preference judgments instead of explicit user feedback \cite{lee2023rlaif, bai2022constitutional, madaan2023selfrefine}. Embedded within an image generation pipeline, our agent effectively turns ordinary inference cycles into opportunities for self-alignment – progressively enhancing semantic accuracy, compositional control, and visual fidelity with each iteration. Over prolonged use, these self-updates accumulate into a form of lifelong learning, analogous to how humans consolidate experience into knowledge \cite{chen2016lifelong, kirkpatrick2017ewc, rebuffi2017icarl, lopez2017gem}. In this way, the system combines modular plug-and-play integration with sustained cognitive growth, offering a concrete path toward self-evolving generative intelligence.

Empirical evaluation shows that our approach yields substantial performance gains on the GenEval, DPG-Bench, and T2I-Bench benchmarks \cite{ganeval2023, dpgbench2024, t2icompbench2023}. On GenEval (object-focused alignment tasks), the agent surpasses prior diffusion model baselines by a significant margin, demonstrating improved compositional grounding and attribute accuracy. On both DPG-Bench and T2I-Bench, our system also achieves higher overall preference satisfaction. Beyond quantitative metrics, we observe an emergent trajectory refinement behavior in qualitative analyses: the learner gradually internalizes the verifier’s reasoning heuristics, leading to progressively more structured generation outputs and higher semantic coherence over successive refinement iterations. Taken together, these results support the hypothesis that the continual learning paradigm – augmented with partial-history replay and dual-scale memory – can serve as a scalable mechanism for sustained intelligence growth in generative models.
\section{Method}
\label{sec:method}
\subsection{Verification and Preference Data Generation}
\label{subsec:verification}
\begin{algorithm}[t]
\caption{\textsc{VerifyAndGenPairs}: Verifier-driven pair construction (Sec.~\ref{subsec:verification})}
\label{alg:verify}
\begin{algorithmic}[1]
\Require prompt $p$, learner $\mathcal{L}_\theta$, verifier $\mathcal{V}$, max steps $T$, threshold $\tau$
\State $\mathcal{C}(p) \gets \mathcal{V}_{\text{formulate}}(p)$ \Comment{Eq.~\ref{eq:1}, produce $\{c_1,\dots,c_n\}$}
\State $\mathcal{X} \gets [\;]$ \Comment{trajectory images}
\For{$t=0$ \textbf{to} $T-1$}
  \If{$t=0$} 
    \State $\mathbf{x}_0 \gets \mathcal{L}_\theta(p)$
  \Else
    \State $\mathbf{x}_t \gets \mathcal{L}_\theta(p, f_{t-1})$ \Comment{Refine with critique, Eq.~\ref{eq:5}}
  \EndIf
  \State $\mathbf{s}^t \gets [\mathcal{V}_{\text{eval}}(c_i,\mathbf{x}_t)]_{i=1}^n$ 
  \If{$\min_i s_i^t \ge \tau$} \Comment{All conditions satisfied}
    \State $\mathbf{x}^+ \gets \mathbf{x}_t$; \; $\mathcal{X}^{-} \gets \{\mathbf{x}_k\}_{k=0}^{t-1}$ \Comment{Negatives are earlier attempts}
    \State \Return $\{(p, \mathbf{x}_k^{-}, \mathbf{x}^+)\}_{k=0}^{t-1}$ 
  \Else
    \State $f_t \gets \mathcal{V}_{\text{critique}}(\mathcal{C}(p), \mathbf{s}^t)$ 
    \State append $\mathbf{x}_t$ to $\mathcal{X}$
  \EndIf
\EndFor
\State \Return $\emptyset$ \Comment{Discard if no positive within $T$}
\end{algorithmic}
\end{algorithm}

The SuperIntelliAgent framework begins with a paired interaction between a \emph{trainable learner} $\mathcal{L}$ (a diffusion model) and a \emph{frozen verifier} $\mathcal{V}$ (a vLLM).  
Given a user prompt $p$, the learner first synthesizes an image $\mathbf{x}_0 = \mathcal{L}(p)$.  
The verifier then performs two reasoning steps: \textbf{condition formulation} and \textbf{alignment evaluation}.

\paragraph{Condition Formulation.}  
\label{condition}
The verifier decomposes the prompt $p$ into a set of semantically grounded sub-goals:
\begin{equation}
\mathcal{C}(p) = \{ c_1, c_2, \dots, c_n \},
\label{eq:1}
\end{equation}

where each $c_i$ denotes a discrete semantic constraint derived from $p$ (e.g., ``object: banana'', ``color: green'').  
This decomposition is obtained through a chain-of-thought–driven consistency audit:
\begin{equation}
\mathcal{C}(p) = \mathcal{V}_{\text{formulate}}(p).
\label{eq:2}
\end{equation}

\paragraph{Alignment Evaluation.}  
For each generated image $\mathbf{x}_t$, the verifier evaluates cross-modal entailment between $\mathbf{x}_t$ and each condition $c_i$:
\begin{equation}
s_i^t = \mathcal{V}_{\text{eval}}(c_i, \mathbf{x}_t) \in [0,1],
\label{eq:3}
\end{equation}
where $s_i^t$ represents the confidence that condition $c_i$ is satisfied.  
The resulting vector $\mathbf{s}^t = [s_1^t, s_2^t, \dots, s_n^t]$ encodes the fine-grained semantic alignment signature between the prompt and generation.

\paragraph{Iterative Verification–Refinement Loop.}  
If all conditions are satisfied ($\min_i s_i^t \ge \tau$), the trajectory terminates with a \emph{positive} sample $\mathbf{x}_t^{+}$.  
Otherwise, the verifier generates structured critique feedback:
\begin{equation}
f_t = \mathcal{V}_{\text{critique}}(\mathcal{C}(p), \mathbf{s}^t),
\end{equation}
and the learner regenerates:
\begin{equation}
\mathbf{x}_{t+1} = \mathcal{L}(p, f_t).
\label{eq:4}
\end{equation}
The loop empirically proceeds for up to $T = 5$ iterations or until all conditions are met.

\paragraph{DPO Pair Construction.}  
If a trajectory yields a valid $\mathbf{x}_T^{+}$, all intermediate generations  
$\{ \mathbf{x}_0, \mathbf{x}_1, \dots, \mathbf{x}_{T-1} \}$  
are labeled as negatives $\{ \mathbf{x}^-_k \}$.  
Each pair $(\mathbf{x}^-_k, \mathbf{x}^+)$ defines a Direct Preference Optimization (DPO) training tuple:
\begin{equation}
\mathcal{D}_{\text{DPO}} = \{ (p, \mathbf{x}^-_k, \mathbf{x}^+) \}.
\label{eq:5}
\end{equation}
If no positive outcome appears within $T$ steps, the trajectory is discarded.  
Pairs from identical prompts are shuffled across mini-batches (batch size $B=16$) to prevent local mode bias.  
This process transforms unsupervised generation–verification cycles into structured preference data, enabling automatic self-supervision without human annotation.


\subsection{Asynchronous Training and DPO Optimization}
\label{sec:sai}
\begin{algorithm}[t]
\caption{SuperIntelliAgent: Asynchronous Training and DPO Optimization \ref{sec:sai}}
\label{alg:superintelliagent}
\begin{algorithmic}[1]
\Require Learner $\mathcal{L}_{\theta}$ (diffusion), frozen verifier $\mathcal{V}$, replay buffer $\mathcal{B}_{\text{replay}}$, batch size $B$, max refine steps $T$, threshold $\tau$, lag bound $K$
\While{system is running}
  \State \textbf{(Inference thread)} sample prompts $\{p_i\}_{i=1}^{B}$
  \ForAll{$p_i$ in the batch} \label{line:outer-for}
    \State $\mathcal{D}_i \gets \textsc{VerifyAndMakePairs}(p_i, \mathcal{L}_\theta, \mathcal{V}, T, \tau)$
    \If{$\mathcal{D}_i \neq \emptyset$}
      \State push $\mathcal{D}_i$ into $\mathcal{B}_{\text{replay}}$ \Comment{Only No$\rightarrow$Yes trajectories}
    \EndIf
  \EndFor
  \State \textbf{(Training thread)} sample $\mathcal{B}_t \subset \mathcal{B}_{\text{replay}}$ \Comment{Shuffled across prompts}
  \If{$\mathcal{B}_t \neq \emptyset$}
    \State $\theta \gets \theta - \eta \, \nabla_\theta \, \mathcal{L}_{\text{DDPO}}(\theta; \mathcal{B}_t)$ 
    \State optionally update LoRA adapters only 
  \EndIf
  \State enforce bounded staleness between generation model and updated $\theta$ ($\le K$ steps)
\EndWhile
\end{algorithmic}
\end{algorithm}

Once the dataset $\mathcal{D}_{\text{DPO}} = \{ (p_i, x_i^-, x_i^+) \}_{i=1}^{N}$ is collected, the learner $\mathcal{L}_\theta$ is optimized using a diffusion-based variant of Direct Preference Optimization (DPO) \cite{rafailov2023dpo}.  

Let $\pi_\theta(x \mid p)$ denote the generative distribution of the learner.  
For each preference pair $(p, x^-, x^+)$, the DPO loss encourages the model to assign higher likelihood to $x^+$:
\begin{equation}
\begin{aligned}
\mathcal{L}_{\text{DPO}}(\theta)
= {}& - \mathbb{E}_{(p,x^-,x^+)} \Big[
  \log \sigma\Big(
  \beta \big[
  \log \pi_\theta(x^+ \mid p) 
\\ &\qquad\qquad\qquad
  - \log \pi_\theta(x^- \mid p)
  \big]\Big)
  \Big].
\end{aligned}
\end{equation}
where $\sigma(\cdot)$ is the logistic sigmoid and $\beta$ controls the separation margin.

For diffusion learners, explicit likelihoods are intractable.  
Following \textsc{DiffusionDPO} \cite{wallace2023diffusiondpo}, we approximate:
\begin{equation}
\log \pi_\theta(x \mid p) \approx -\mathcal{L}_{\text{denoise}}(\theta; p, x) + \text{const},
\end{equation}
where $\mathcal{L}_{\text{denoise}}$ is the standard diffusion denoising loss.  
Hence, the optimization simplifies to:
\begin{equation}
\mathcal{L}_{\text{DDPO}}(\theta)
= \mathbb{E}_{(p,x^-,x^+)} 
  \big[
    \mathcal{L}_{\text{denoise}}(\theta; p, x^+) 
  - \mathcal{L}_{\text{denoise}}(\theta; p, x^-)
  \big].
\end{equation}

\paragraph{Asynchronous Pipeline.}  
SuperIntelliAgent decouples inference and training into parallel threads.  
At iteration $t$:
\begin{enumerate}
    \item The learner $\mathcal{L}_{\theta_t}$ generates a mini-batch of $B$ samples for multiple prompts.
    \item The verifier produces new labeled pairs $\Delta\mathcal{D}_t$, which are appended to the replay buffer.
    \item Simultaneously, the learner samples a batch $\mathcal{B}_t \subset \text{buffer}$ and performs gradient updates on $\mathcal{L}_{\text{DDPO}}$.
    \item Parameters are updated to $\theta_{t+1}$, and subsequent generations use the new model weights.
\end{enumerate}
This asynchronous loop ensures that training and inference co-exist, enabling continual online adaptation during deployment while maintaining stability via a bounded lag $K$ between generation and parameter updates.


\subsection{Diffusion Model as Learner}

Our learner $\mathcal{L}_\theta$ is a denoising diffusion probabilistic model (DDPM) \cite{ho2020denoising} adapted for preference optimization.  
Given a clean image $\mathbf{x}_0$ drawn from the data distribution, the forward process gradually adds Gaussian noise:
\begin{equation}
q(\mathbf{x}_t \mid \mathbf{x}_0) = \mathcal{N}(\sqrt{\bar{\alpha}_t}\mathbf{x}_0, (1-\bar{\alpha}_t)\mathbf{I}),
\end{equation}
where $\bar{\alpha}_t = \prod_{s=1}^t (1-\beta_s)$ and $\beta_s$ is a variance schedule.  
The reverse process learns to denoise step by step:
\begin{equation}
p_\theta(\mathbf{x}_{t-1} \mid \mathbf{x}_t, p) = 
\mathcal{N}\!\big(
\mathbf{x}_{t-1}; 
\mu_\theta(\mathbf{x}_t, t, p), 
\sigma_t^2 \mathbf{I}
\big),
\end{equation}
where $\mu_\theta$ is predicted from a U-Net backbone conditioned on the prompt embedding.

During DPO fine-tuning, $\mathcal{L}_{\text{denoise}}$ in Eq.~(8) serves as the differentiable proxy for preference likelihood.  
For each sample $(p, x^-, x^+)$, we minimize:
\begin{equation}
\mathcal{L}_{\text{DDPO}} = 
\mathbb{E}_t \big[
\| \epsilon_\theta(\mathbf{x}_t^+, t, p) - \epsilon \|^2
- 
\| \epsilon_\theta(\mathbf{x}_t^-, t, p) - \epsilon \|^2
\big],
\end{equation}
where $\epsilon$ denotes the noise sampled during diffusion training.  
This aligns the model to prefer outputs that better satisfy the verifier’s criteria while maintaining diffusion stability.

Unlike reinforcement-learning–based alignment methods, our diffusion-DPO formulation yields stable gradients and enables smooth interpolation between preference manifolds. This stability allows the learner to evolve continuously without collapsing into discrete preference modes—an essential property for sustaining open-ended generative intelligence.


\subsection{Continual Adaptation and LoRA-based Learning}

SuperIntelliAgent aims for continual intelligence growth—adapting persistently as new experience is gathered.  
Whenever new preference data are generated, the learner can trigger fine-tuning without human intervention.  
To maintain efficiency and prevent catastrophic forgetting, we adopt a replay-buffer strategy:  
only trajectories exhibiting most measurable progress are stored for future sampling.  
Formally, the replay buffer $\mathcal{B}_{\text{replay}}$ is defined as:
\begin{equation}
\mathcal{B}_{\text{replay}} = \{ (p, \mathbf{x}^-_k, \mathbf{x}^+) \mid \exists \, k : \text{progress}(\mathbf{x}^-_k \!\rightarrow\! \mathbf{x}^+) = 1 \}.
\end{equation}
During continual updates, $\mathcal{B}_{\text{replay}}$ serves as an auxiliary supervision source to reinforce stable learning while bootstrapping increasingly complex curricula.

\paragraph{Parameter-Efficient Fine-tuning via LoRA.}  
To support frequent online updates with minimal computational cost, the learner employs Low-Rank Adaptation (LoRA) \cite{hu2021lora}.  
Given a frozen weight matrix $\mathbf{W} \in \mathbb{R}^{d \times k}$, LoRA introduces two low-rank matrices $\mathbf{A} \in \mathbb{R}^{d \times r}$ and $\mathbf{B} \in \mathbb{R}^{r \times k}$ such that:
\begin{equation}
\mathbf{W}' = \mathbf{W} + \Delta \mathbf{W}, \quad
\Delta \mathbf{W} = \mathbf{A} \mathbf{B}^{\top}, \quad r \ll \min(d,k).
\end{equation}
Only $\mathbf{A}$ and $\mathbf{B}$ are trainable, while $\mathbf{W}$ remains frozen.  
This design allows the diffusion model to adapt dynamically to new preference data with negligible overhead, enabling on-the-fly fine-tuning even during inference.

\paragraph{Online Lifelong Learning.}  
By combining LoRA-based updates with asynchronous training, SuperIntelliAgent achieves genuine online continual learning.  
New data samples—generated during normal inference—can immediately participate in model updates without pausing the system.  
As a result, the learner progressively internalizes the verifier’s reasoning heuristics, consolidating them into its parameters as long-term memory.  
Over time, this mechanism transforms continuous usage into cumulative intelligence growth, analogous to human lifelong learning, where experience is perpetually assimilated into knowledge.

\subsection{Summary}

In essence, SuperIntelliAgent unifies reasoning, generation, and continual adaptation into a single closed-loop learning architecture.  
A frozen verifier transforms textual intent into structured semantic conditions, converting free-form prompts into verifiable learning objectives.  
A trainable diffusion learner responds to these objectives, refining its generations through iterative feedback until all semantic conditions are satisfied.  
This dynamic yields not only labeled preference data for Direct Preference Optimization, but also a replayable trajectory of progressive reasoning steps—forming the backbone of continual learning.  
By coupling asynchronous training with parameter-efficient LoRA adaptation, SuperIntelliAgent turns each inference episode into an opportunity for internal refinement.  
Over time, the system continuously consolidates these micro-updates into long-term representational shifts, mirroring how humans accumulate knowledge through repeated interaction and reflection.  
Through this synergy of autonomous supervision, diffusion-based alignment, and lifelong adaptation, SuperIntelliAgent represents a practical step toward generative agents that not only perform but also evolve.
\section{Experiments}
\label{sec:experiments}

We empirically evaluate SuperIntelliAgent as a continual, self-improving diffusion learner on three challenging text-to-image benchmarks: \textbf{GenEval}, \textbf{DPG-Bench}, and \textbf{T2I-Bench}. Note that SuperIntelliAgent does not require a training dataset; it only needs a single pass over the evaluation dataset.

\subsection{Benchmarks and Metrics}

\paragraph{GenEval}
GenEval is a VQA-style benchmark designed to measure fine-grained compositional alignment between text prompts and generated images. Each prompt is evaluated by an image--text question answering model, and the benchmark is organized into six categories: \texttt{color\_attr}, \texttt{counting}, \texttt{colors}, \texttt{position}, \texttt{single\_object}, and \texttt{two\_object}. Following prior work, we report per-category accuracy and overall accuracy over 553 prompts.

\paragraph{DPG-Bench.}
DPG-Bench focuses on diffusion preference generalization with detailed compositional prompts. For each prompt we score it using a BLIP-based VQA model. We report the average DPG score (0--1, shown as a percentage) over 1{,}065 prompts.

\paragraph{T2I-CompBench.}
T2I-CompBench is a large-scale benchmark for compositional text-to-image generation, covering attribute binding and relational reasoning. This dataset includes eight categories: \texttt{color\_val}, \texttt{shape\_val}, \texttt{texture\_val}, \texttt{spatial\_val}, \texttt{3d\_spatial\_val}, \texttt{numeracy\_val}, \texttt{non\_spatial\_val}, and \texttt{complex\_val}, and we report the BLIP-VQA score averaged over 640 prompts.

\subsection{Models and Training Protocol}

\paragraph{Learner backbones.}
We instantiate the trainable diffusion learner with three vision--language diffusion models:
(1) \textbf{Stable Diffusion-v1.5}, a small vLLM 0.98B-parameter model, and
(2) \textbf{Janus-1.3B}, a compact 1.3B-parameter model, and
(3) \textbf{Janus-Pro-7B}, a stronger 7B-parameter variant.
Unless otherwise noted, all models are fine-tuned with LoRA-style adapters.

\paragraph{Verifier setup.}
The frozen verifier side of SuperIntelliAgent is made up of:
a \emph{judge} (GPT-4o-mini) that assigns a scalar score in $[0,100]$ plus a textual rationale to each generated sample, and an \emph{improver} (o1-mini / GPT-4o-mini) that proposes refined prompts or guidance when the score is low. The learner and verifier interact in short threads: for a given prompt, the learner generates an image, the judge evaluates it, and if the score falls below a threshold the improver synthesizes a better-targeted prompt before returning critique feedback and regeneration.

\paragraph{Auto-DPO pair construction.}
Our continual Auto-DPO loop automatically converts the above interactions into preference data. For each prompt:
(1) if the first sample already achieves a high score, the prompt is skipped (no training signal);
(2) otherwise, the original lower-scoring image (or any intermediate higher-scoring revision) paired with the final accepted output yields multiple \emph{rejected}–\emph{chosen} preference pairs. We retain only those pairs whose score difference exceeds a minimum margin (0.15 in normalized units), ensuring that DPO updates are driven by confidently separated preferences.

\paragraph{Continual training schedule.}
We process benchmarks in a streaming fashion. Prompts are fed through the learner--verifier loop, and DPO pairs accumulate in a buffer. After every $N$ prompts (Table~\ref{tab:hyperparams}), We perform a short fine-tuning burst of $K$ optimization steps on the accumulated preference pairs, after which generation resumes with the updated learner.
This update can run synchronously—briefly pausing generation—or asynchronously, in which case generation continues with the previous learner while its parameters are being updated in the background. This realizes continual, on-the-fly fine-tuning rather than offline batch training. We use DPO with $\beta = 0.5$ for all experiments, a batch size of $2$, and a single CUDA device. Because the replay buffer is small relative to the model size, individual batches may be sampled multiple times across successive training steps.

\begin{table*}[t]
\centering
\caption{Hyperparameters for continual DPO. \texttt{batch} is the $N$ prompts number between fine-tuning bursts.}
\label{tab:hyperparams}
\begin{tabular}{lcccc}
\toprule
Benchmark & Model & LR & Train freq. & Steps \\
\midrule
GenEval & Stable Diffusion-v1.5 & $1{\times}10^{-6}$ & 32 & 8 \\
GenEval & Janus-1.3B & $7{\times}10^{-5}$ & 32 & 8 \\
GenEval & Janus-Pro-7B & $5{\times}10^{-5}$ & 32 & 16 \\
DPG-Bench & Janus-1.3B & $5{\times}10^{-5}$ & 16 & 8 \\
DPG-Bench & Janus-Pro-7B & $3{\times}10^{-5}$ & 16 & 8 \\
T2I-CompBench & Janus-1.3B & $7{\times}10^{-5}$ & 32 & 8 \\
T2I-CompBench & Janus-Pro-7B & $3{\times}10^{-5}$ & 16 & 8 \\
\bottomrule
\end{tabular}
\end{table*}

\subsection{Overall Quantitative Results}

Table~\ref{tab:main_results} summarizes baseline (no continual learning) and SuperIntelliAgent-trained performance for both learner sizes across all three benchmarks.

\begin{table*}[t]
\centering
\caption{Overall performance (\%) of Janus-1.3B and Janus-Pro-7B before (Baseline) and after continual Auto-DPO fine-tuning with SuperIntelliAgent.}
\label{tab:main_results}
\begin{tabular}{lcccc}
\toprule
& \multicolumn{2}{c}{Janus-1.3B} & \multicolumn{2}{c}{Janus-Pro-7B} \\
Benchmark & Baseline & SuperIntelliAgent & Baseline & SuperIntelliAgent \\
\midrule
GenEval & 58.41 & \textbf{69.62} & 76.31 & \textbf{83.54} \\
DPG-Bench & 83.09 & \textbf{84.57} & 87.13 & \textbf{88.35} \\
T2I-CompBench & 52.43 & \textbf{54.70} & 60.61 & \textbf{62.09} \\
\bottomrule
\end{tabular}
\end{table*}

\paragraph{Benchmarks.}
On GenEval, SuperIntelliAgent improves Janus-1.3B from $58.41\%$ to $69.62\%$ accuracy and Janus-Pro-7B from $76.31\%$ to $83.54\%$. The larger backbone maintains a substantial advantage: after fine-tuning, Janus-Pro-7B still outperforms Janus-1.3B by +13.92 points (83.54\% vs.\ 69.62\%). This indicates that the proposed continual learning complements, rather than replaces, scaling of the underlying generative model.
On DPG-Bench, SuperIntelliAgent yields gains for Janus-1.3B (+1.48) and Janus-Pro-7B (+1.24), reaching 88.35\% DPG score. T2I-CompBench is the hardest benchmark, improvements from continual learning are modest (+2.27 and +1.48 points, respectively). 

\paragraph{Relative benchmark difficulty.}
Ranking by performance, GenEval is easiest (69.62/83.54\%), and T2I-Bench is clearly the most challenging (54.70/62.09). This ordering matches qualitative impressions: GenEval prompts are detailed but structurally regular, whereas T2I-CompBench stresses compositional generalization and numeracy.

\subsection{GenEval Per-Category Analysis}

To understand where SuperIntelliAgent helps, we break down GenEval into its six categories.

\paragraph{Comparison across baseline small generative models}
Table~\ref{tab:gen_eval_all} compares Stable Diffusion v1.5 and v2.1 against our
\textbf{SuperIntelliAgent + Stable Diffusion v1.5} variant on the GenEval benchmark.
Despite using the weaker backbone (v1.5), SuperIntelliAgent substantially improves
compositional correctness and achieves an \textbf{overall score of 0.525}, 
surpassing both SD~v1.5 (0.06) and even SD~v2.1 (0.17). 
This gain is driven by several factors. First, the method nearly perfects
single-object recognition (0.99) and significantly boosts counting accuracy
(0.61 vs.\ 0.04/0.07), showing that the proposed continual learning is especially
effective at correcting numeracy and object-quantity failures.
Second, two-object relations improve from 0.76 (v1.5) to 0.55 in our method,
while color and appearance attributes remain competitive with SD~v2.1.
Although gains on position and color\_attr remain modest—reflecting persistent
difficulty in spatial grounding—the overall improvement demonstrates that
SuperIntelliAgent provides a strong self-supervised alignment signal that
compensates for weaknesses of the underlying diffusion model. Remarkably,
the resulting model not only outperforms its original backbone but also exceeds
the overall GenEval score of Stable Diffusion v2.1, despite relying on SD~v1.5
as the base generator.

\begin{table*}[t]
\centering
\small
\begin{tabular}{lccccc}
\toprule
\textbf{Category} 
& \textbf{CLIP Ret.} 
& \textbf{minDALL-E} 
& \textbf{SD v1.5} 
& \textbf{SD v2.1} 
& \textbf{SAI + SD v1.5} \\
\midrule
Overall            & 0.35 & 0.23 & 0.43 & 0.50 & \textbf{0.53} \\
Single object      & 0.89 & 0.73 & 0.97 & 0.98 & \textbf{0.99} \\
Two object         & 0.22 & 0.11 & 0.38 & 0.51 & \textbf{0.55} \\
Counting           & 0.37 & 0.12 & 0.35 & 0.44 & \textbf{0.61} \\
Colors             & 0.62 & 0.37 & 0.76 & 0.85 & \textbf{0.79} \\
Position           & 0.03 & 0.02 & 0.04 & 0.07 & \textbf{0.11} \\
Color attribution  & 0.00 & 0.01 & 0.06 & 0.17 & \textbf{0.11} \\
\bottomrule
\end{tabular}
\caption{GenEval category-wise comparison across baseline small generative models, Stable Diffusion variants, and our \textbf{SuperIntelliAgent + SD v1.5}.}
\label{tab:gen_eval_all}
\end{table*}

\paragraph{Janus-1.3B.}
Table~\ref{tab:ganeval_janus13b} shows that continual learning particularly benefits difficult compositional categories. Counting accuracy nearly doubles, from $23.75\%$ to $46.25\%$ (+22.50), and two-object compositions improve from $60.61\%$ to $84.85\%$ (+24.24). Position and color-attribute binding also see sizable gains. Single-object prompts are already near-perfect and remain unchanged.

\begin{table}[t]
\centering
\caption{GenEval per-category accuracy (\%) for Janus-1.3B.}
\label{tab:ganeval_janus13b}
\begin{tabular}{lccc}
\toprule
Category & Baseline & SuperIntelliAgent & $\Delta$ \\
\midrule
color\_attr    & 43.00 & 52.00 & +9.00 \\
counting       & 23.75 & 46.25 & +22.50 \\
colors         & 81.91 & 86.17 & +4.26 \\
position       & 45.00 & 52.00 & +7.00 \\
single\_object & 98.75 & 98.75 & +0.00 \\
two\_object    & 60.61 & 84.85 & +24.24 \\
\midrule
Overall        & 58.41 & 69.62 & +11.21 \\
\bottomrule
\end{tabular}
\end{table}

\paragraph{Janus-Pro-7B.}
The larger learner shows similar trends at a higher absolute level (Table~\ref{tab:ganeval_januspro}). Counting rises from $55.00\%$ to $71.25\%$ (+16.25), position from $75.00\%$ to $81.00\%$ (+6.00), and color-attribute binding from $60.00\%$ to $69.00\%$ (+9.00). Two-object prompts reach $92.93\%$ accuracy. Taken together, these results indicate that the continual learning is particularly effective at tightening complex compositional constraints and alleviating counting failures.

\begin{table}[t]
\centering
\caption{GenEval per-category accuracy (\%) for Janus-Pro-7B.}
\label{tab:ganeval_januspro}
\begin{tabular}{lccc}
\toprule
Category & Baseline & SuperIntelliAgent & $\Delta$ \\
\midrule
color\_attr    & 60.00 & 69.00 & +9.00 \\
counting       & 55.00 & 71.25 & +16.25 \\
colors         & 87.23 & 89.36 & +2.13 \\
position       & 75.00 & 81.00 & +6.00 \\
single\_object & 98.75 & 98.75 & +0.00 \\
two\_object    & 82.83 & 92.93 & +10.10 \\
\midrule
Overall        & 76.31 & 83.54 & +7.23 \\
\bottomrule
\end{tabular}
\end{table}
\paragraph{Qualitative evaluation}

\begin{figure*}[t]
    \centering

    \begin{subfigure}{0.19\linewidth}
        \centering
        \includegraphics[width=\linewidth]{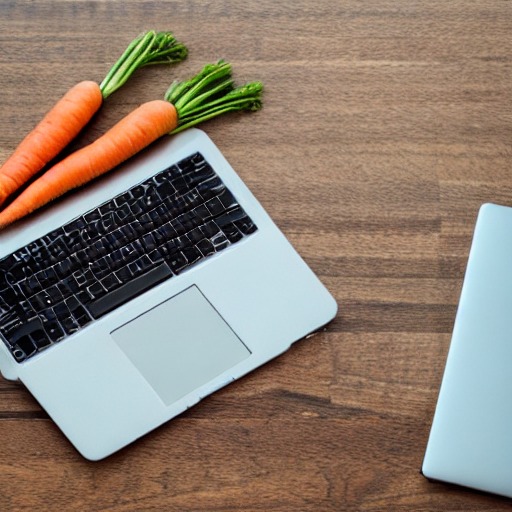}
        \caption*{Laptop+Carrot (Base)}
    \end{subfigure}
    \begin{subfigure}{0.19\linewidth}
        \centering
        \includegraphics[width=\linewidth]{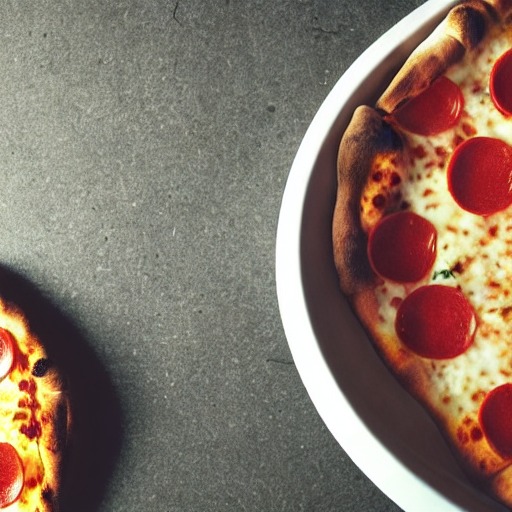}
        \caption*{Bowl+Pizza (Base)}
    \end{subfigure}
    \begin{subfigure}{0.19\linewidth}
        \centering
        \includegraphics[width=\linewidth]{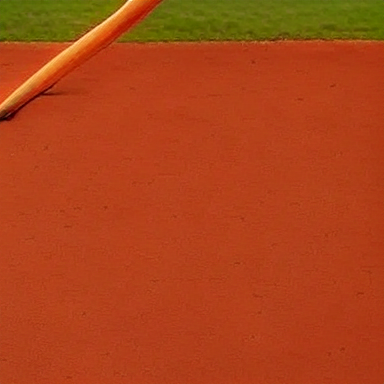}
        \caption*{Bat (Base)}
    \end{subfigure}
    \begin{subfigure}{0.19\linewidth}
        \centering
        \includegraphics[width=\linewidth]{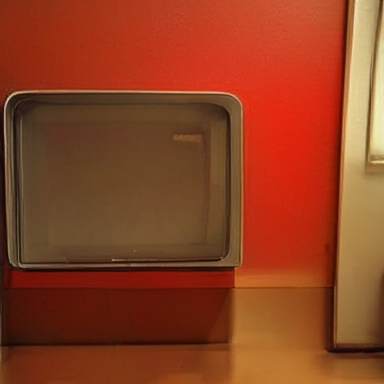}
        \caption*{Microwave (Base)}
    \end{subfigure}
    \begin{subfigure}{0.19\linewidth}
        \centering
        \includegraphics[width=\linewidth]{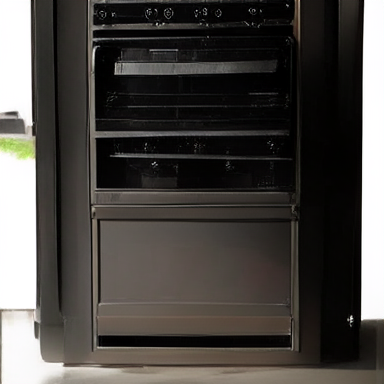}
        \caption*{Oven (Base)}
    \end{subfigure}
    \hfill
    \begin{subfigure}{0.19\linewidth}
        \centering
        \includegraphics[width=\linewidth]{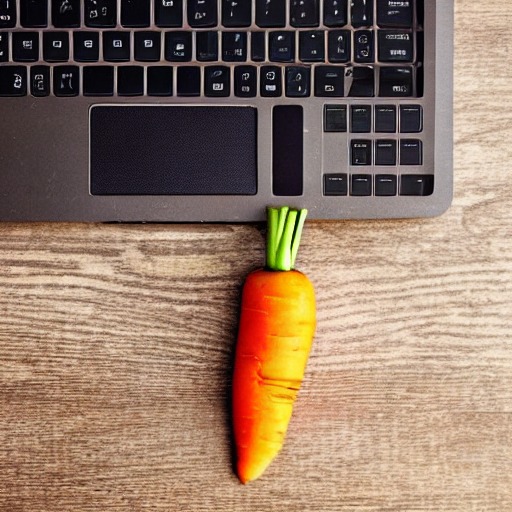}
        \caption*{Laptop+Carrot (Trained)}
    \end{subfigure}
    \begin{subfigure}{0.19\linewidth}
        \centering
        \includegraphics[width=\linewidth]{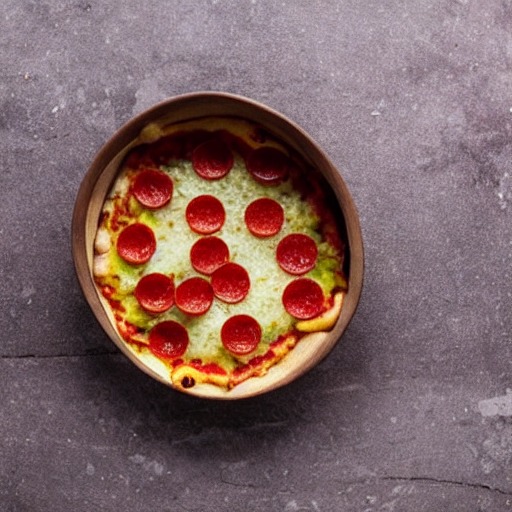}
        \caption*{Bowl+Pizza (Trained)}
    \end{subfigure}
    \begin{subfigure}{0.19\linewidth}
        \centering
        \includegraphics[width=\linewidth]{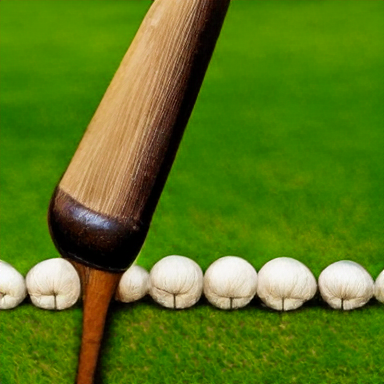}
        \caption*{Bat (Trained)}
    \end{subfigure}
    \begin{subfigure}{0.19\linewidth}
        \centering
        \includegraphics[width=\linewidth]{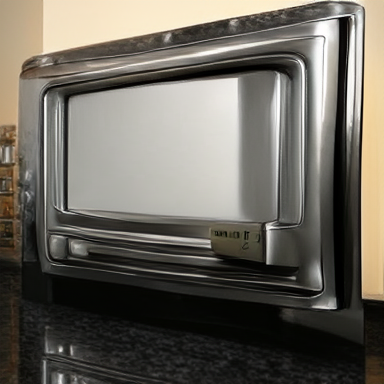}
        \caption*{Microwave (Trained)}
    \end{subfigure}
    \begin{subfigure}{0.19\linewidth}
        \centering
        \includegraphics[width=\linewidth]{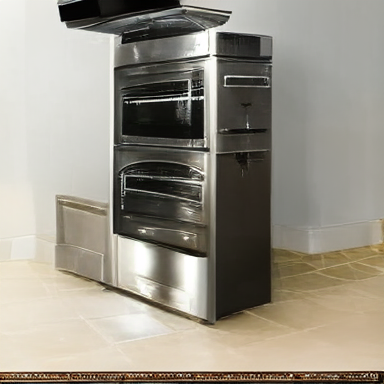}
        \caption*{Oven (Trained)}
    \end{subfigure}

    \caption{Qualitative comparisons between baseline Janus outputs and images produced after continual training with SuperintelliAgent across five GenEval prompts.}
    \label{fig:qualitative}
\end{figure*}

We further evaluate qualitative improvements introduced by SuperintelliAgent on several GenEval prompts. Fig.~\ref{fig:qualitative} presents the \emph{baseline Janus outputs} (negative) alongside the corresponding \emph{SuperIntelliAgent-trained outputs} (positive). Across all examples, the trained model consistently restores missing object relationships and enforces coherent spatial layouts. For the prompt \emph{a photo of a laptop and a carrot}, the baseline often merges the carrot into the keyboard, distorts key geometry, or hallucinates multiple carrots, whereas the trained version clearly depicts a single carrot and a laptop with proper separation and intact structure. For \emph{a bowl and a pizza}, the baseline places the pizza partly outside the bowl, violating the relational constraint; SuperintelliAgent correctly positions the pizza inside the bowl with clean boundaries and well-formed pepperoni slices. Similar gains appear in object-identification prompts: the baseline \emph{baseball bat} resembles a featureless wooden rod lacking grip and taper, while the trained model produces a realistic bat with appropriate material cues. For appliances, the baseline \emph{microwave} is frequently misclassified as a CRT-style screen, and the baseline \emph{oven} shows broken or incomplete geometry. After continual learning, the corrected outputs display recognizable structural elements—door seams, control panels, symmetric cavities—indicating stronger category understanding. Overall, these qualitative comparisons demonstrate that SuperintelliAgent reduces hallucinations, repairs local texture and attribute errors, and internalizes higher-level compositional constraints, resulting in more faithful, semantically aligned multi-object generations.
\paragraph{Scaling comparison.}
Comparing the trained models, Janus-Pro-7B delivers large absolute gains over Janus-1.3B on the hardest categories: +25.00 points on counting (71.25\% vs.\ 46.25\%), +29.00 on position (81.00\% vs.\ 52.00\%), and +17.00 on color-attributes (69.00\% vs.\ 52.00\%). This suggests that a stronger learner can more fully internalize the verifier’s structured feedback, especially on compositional and relational aspects.

\subsection{Training Efficiency and Supervision Density}

An advantage of SuperIntelliAgent is that it only trains on \emph{hard} cases where the verifier finds room for improvement.

\paragraph{DPG-Bench.}
On DPG-Bench with Janus-1.3B, the system processes 1{,}065 prompts but generates only 241 DPO pairs, skipping 1033 prompts whose initial generations already score above threshold, or finally cannot generate positive samples after iterations(note that one prompt can generate multiple pairs due to the existing intermediate generations, and only 32 data are used for fine-tuning). This corresponds to only approximately $3\%$ of prompts contributing to fine-tuning, yet overall accuracy still improves from 83.09\% to 84.57\%. Note that all data are consumed in a single pass: each sample is first inferenced, then used for fine-tuning, and never re-inferenced again (though it may reappear in the replay buffer for additional fine-tuning). Similar efficiency is observed for Janus-Pro-7B (168 DPO pairs), where a relatively small number of fine-tuning bursts yields a +1.24 point gain.

\paragraph{T2I-Bench.}
For T2I-Bench, Janus-1.3B with 499 DPO pairs produces 640 prompts, requiring 29 fine-tuning sessions; In comparison, for Janus-7B, 382 DPO pairs are produced. With only approximately ~4\% of data being used for fine-tuning, the final accuracy reaches 54.70\% and 62.09\% respectively.

\paragraph{GenEval.}
For GenEval, Janus-1.3B generates 380 DPO pairs across 553 prompts (10 sessions), and Janus-Pro-7B generates 213 pairs (6 sessions). The larger model requires fewer pairs and fewer sessions to reach a stronger final performance, reflecting better sample efficiency of the learner at larger scale.

\subsection{Discussion}

Overall, the experiments support three key conclusions:

\begin{itemize}
    \item \textbf{Continual self-training works.} Across all benchmarks and all model sizes, the proposed SuperIntelliagent yields consistent improvements over frozen baselines, with especially strong gains on GenEval compositional tasks.
    \item \textbf{Scaling and pairing are complementary.} The larger Janus-Pro-7B learner starts from significantly stronger baselines and still benefits from continual learning. After fine-tuning, it dominates the smaller model on every benchmark, and particularly on counting and spatial reasoning.
    \item \textbf{Benchmarks expose different failure modes.} DPG-Bench appears relatively easy; many prompts are “already good” and do not trigger fine-tuning. GenEval exposes clear weaknesses in counting and object relations that SuperIntelliAgent mitigates. T2I-CompBench remains challenging even after aggressive fine-tuning, indicating that attribute binding and complex multi-object compositions are far from solved.
\end{itemize}

These results suggest that the SuperIntelliAgent demonstrates the potential of continual, self-improving learning, and that such continual learning will likely become a central direction for future text-to-image systems.

\section{Conclusion}
\label{sec:conclusion}

We presented \textbf{SuperIntelliAgent}, a unified framework for continual self-improvement that couples a trainable diffusion learner with a frozen reasoning-based verifier.  
Unlike static foundation models that remain fixed after training, SuperIntelliAgent transforms the inference pipeline itself into a continuous learning loop—where generation, verification, and preference optimization coexist asynchronously.  
Through the proposed Auto-DPO mechanism and selective replay of verifiable progress, the system autonomously constructs its own training data, bootstraps adaptive curricula, and continually consolidates experience into long-term model parameters via lightweight LoRA updates.

Empirical results on GenEval and DPG-Bench demonstrate that this design yields superior semantic alignment, compositional fidelity, and preference consistency over existing baselines.  
Beyond quantitative gains, SuperIntelliAgent offers a new paradigm for scalable, lifelong learning—mirroring how human intelligence evolves through repeated reflection and self-correction.  
Its infrastructure-agnostic formulation enables seamless integration into agentic ecosystems (e.g., Autogen, Semantic Kernel, etc.), allowing deployed systems to accumulate knowledge over time and transform everyday inference cycles into sustained cognitive growth.  
We hope this work inspires further exploration of LLM agent as the minimal reliable unit of emergent, continually improving intelligence.

{
    \small
    \bibliographystyle{ieeenat_fullname}
    \bibliography{main}
}
\newpage
\section{Appendix}
\subsection{Qualitative performance on T2I and DPG Bench}
\begin{figure*}[t]
    \centering

    \begin{subfigure}{0.19\linewidth}
        \centering
        \includegraphics[width=\linewidth]{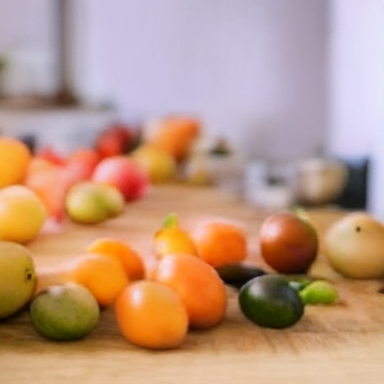}
        \caption*{}
    \end{subfigure}
    \begin{subfigure}{0.19\linewidth}
        \centering
        \includegraphics[width=\linewidth]{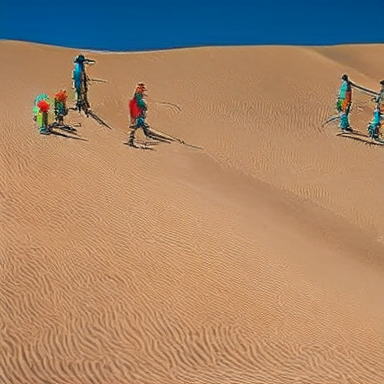}
        \caption*{}
    \end{subfigure}
    \begin{subfigure}{0.19\linewidth}
        \centering
        \includegraphics[width=\linewidth]{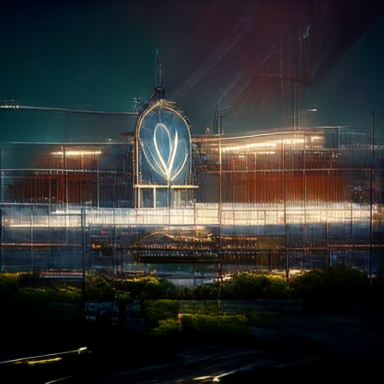}
        \caption*{}
    \end{subfigure}
    \begin{subfigure}{0.19\linewidth}
        \centering
        \includegraphics[width=\linewidth]{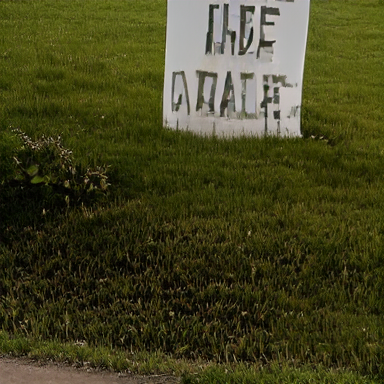}
        \caption*{}
    \end{subfigure}
    \begin{subfigure}{0.19\linewidth}
        \centering
        \includegraphics[width=\linewidth]{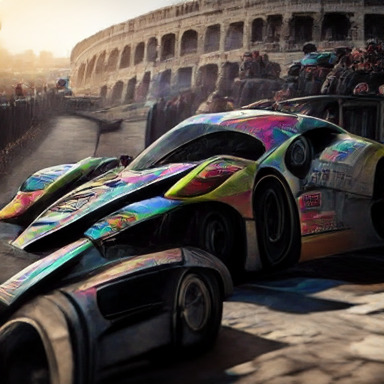}
        \caption*{}
    \end{subfigure}
    \vspace{-6pt} 
    \begin{subfigure}{0.19\linewidth}
        \centering
        \includegraphics[width=\linewidth]{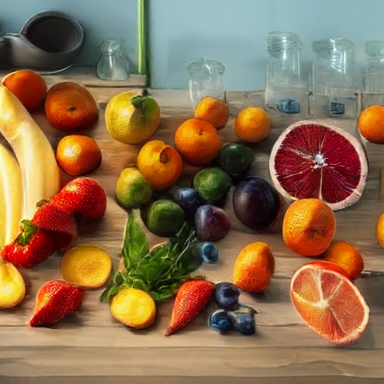}
        \caption*{}
    \end{subfigure}
    \begin{subfigure}{0.19\linewidth}
        \centering
        \includegraphics[width=\linewidth]{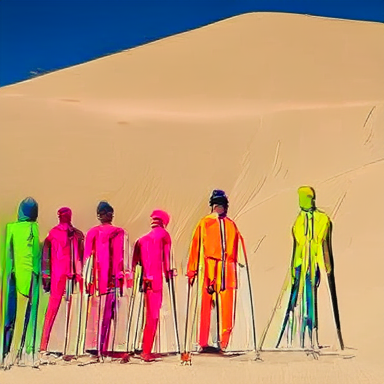}
        \caption*{}
    \end{subfigure}
    \begin{subfigure}{0.19\linewidth}
        \centering
        \includegraphics[width=\linewidth]{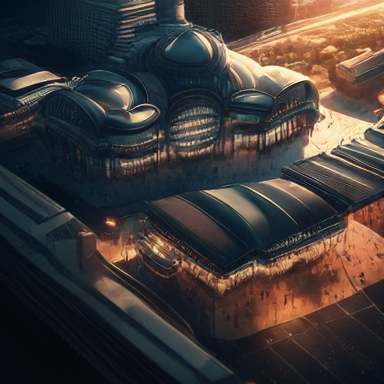}
        \caption*{}
    \end{subfigure}
    \begin{subfigure}{0.19\linewidth}
        \centering
        \includegraphics[width=\linewidth]{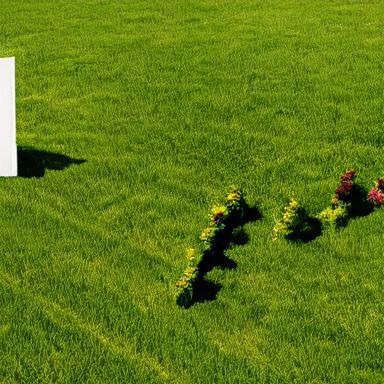}
        \caption*{}
    \end{subfigure}
    \begin{subfigure}{0.19\linewidth}
        \centering
        \includegraphics[width=\linewidth]{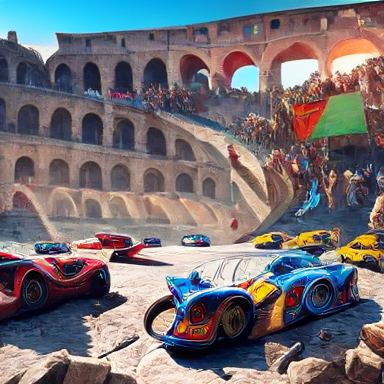}
        \caption*{}
    \end{subfigure}

    \caption{Qualitative comparisons between baseline Janus outputs and images produced after continual training with SuperintelliAgent across five DPG prompts. Column 1: In the foreground of the image, a variety of colorful fruits are scattered across a wooden table, with their fine details and textures in sharp focus. The background features a blurred arrangement of kitchenware and a pastel-colored wall, providing a soft contrast to the vivid sharpness of the fruits on the table. The diffused light gently illuminates the scene, highlighting the smooth skins of the fruits and casting subtle shadows upon the wooden surface; Column 2: a collection of individuals clad in bright ski gear against the contrasting backdrop of a vast beige sand dune. Each person is equipped with skis and poles, ready to ascend the gentle slope of the dune under a clear blue sky. Their colorful attire stands out vividly against the monochrome landscape of sand. Column 3: An extraordinary rendition of Melbourne's Southern Cross Station presented from a bird's-eye view, encapsulated by the signature aesthetics akin to the works of Makoto Shinkai. The image boasts a resolution of 8K, delivering an ultra-detailed and sharply defined portrayal that captures even the subtlest of features. The station and its surroundings are bathed in epic lighting that casts dramatic shadows and projects vivid light refractions across the scene, offering a sense of hyperrealism that's enhanced by the ultra uplight effect. Each element within the composition is rendered with high fidelity, giving life to a photorealistic scene that is both captivating and intricately depicted. Column 4:  A bold, white sign with the words 'KEEP OFF THE GRASS' stands prominently next to a lush, green lawn. The sign, with its stark black lettering, is mounted on a metal pole and positioned at the edge of the neatly trimmed grass. Surrounding the lawn are small flowering plants, adding a touch of color to the scene. Column 5: A dynamic scene unfolds at the historic Colosseum, where a fleet of sleek, multicolored racing cars roar past an excited crowd. The vehicles, adorned with vibrant decals and sponsor logos, navigate a temporary circuit that has been meticulously laid out within the ancient arena's interior. Spectators are perched on stone seats that have withstood the test of time, their attention fixed on the blur of machines vying for the lead under the bright afternoon sun.}
    \label{fig:qualitative_dpg}
\end{figure*}

\begin{figure}[t]
    \centering

    \begin{subfigure}{0.3\linewidth}
        \centering
        \includegraphics[width=\linewidth]{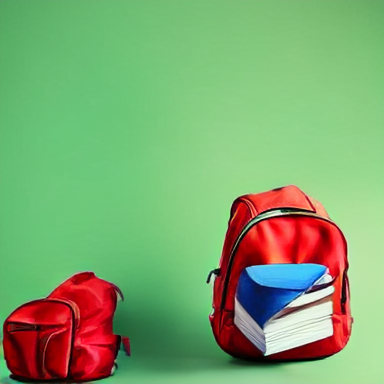}
        \caption*{}
    \end{subfigure}
    \begin{subfigure}{0.3\linewidth}
        \centering
        \includegraphics[width=\linewidth]{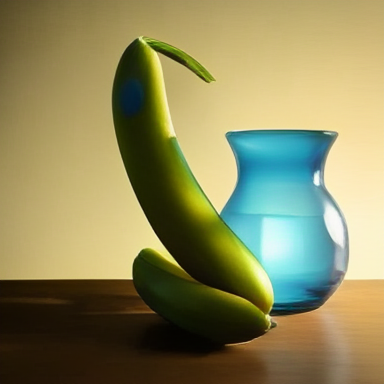}
        \caption*{}
    \end{subfigure}
    \begin{subfigure}{0.3\linewidth}
        \centering
        \includegraphics[width=\linewidth]{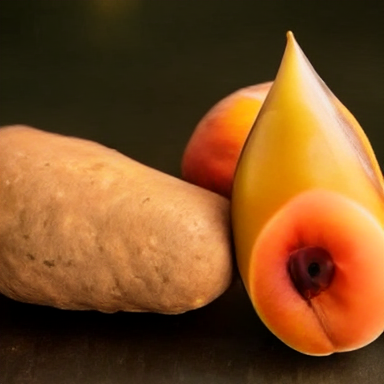}
        \caption*{}
    \end{subfigure}

    \begin{subfigure}{0.3\linewidth}
        \centering
        \includegraphics[width=\linewidth]{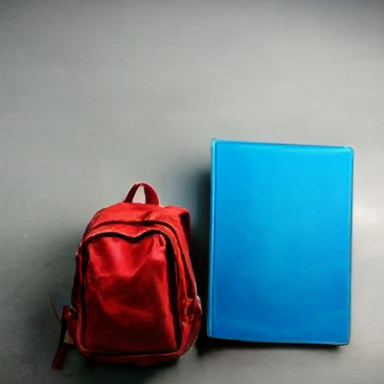}
        \caption*{}
    \end{subfigure}
    \begin{subfigure}{0.3\linewidth}
        \centering
        \includegraphics[width=\linewidth]{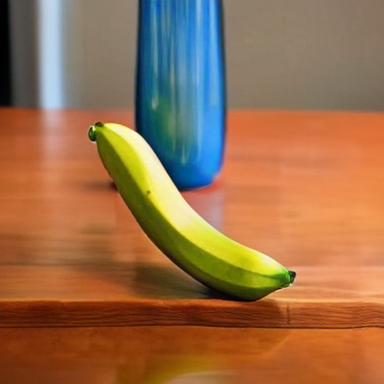}
        \caption*{}
    \end{subfigure}
    \begin{subfigure}{0.3\linewidth}
        \centering
        \includegraphics[width=\linewidth]{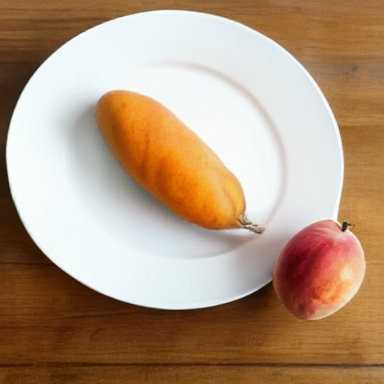}
        \caption*{}
    \end{subfigure}

\caption{Qualitative comparisons between baseline Janus outputs and images produced after continual training with SuperintelliAgent across three T2I prompts. Prompt 1: a red backpack and a blue book; Prompt 2: a green banana and a blue vase; Prompt 3: an oblong sweet potato and a teardrop peach.}
\label{fig:qualitative_t2i}
\end{figure}

More detailed comparisons can be found in figure \ref{fig:qualitative_dpg} and figure \ref{fig:qualitative_t2i}.

\subsection{Representative Prompts Used in SuperIntelliAgent}
\label{appendix:prompts}

SuperIntelliAgent relies on carefully structured prompts for both the learner and verifier, designed to maximize interpretability and reasoning consistency.  
We categorize all prompts into three functional groups: \textbf{generation prompts}, \textbf{verification prompts}, and \textbf{condition generation prompts}.  
Below, we detail their templates and rationale.
\paragraph{Condition Generation Prompts.}
The verifier $\mathcal{V}$ uses chain-of-thought prompting to transform a text prompt into a set of explicit verification conditions $\mathcal{C}(p) = \{c_i\}_{i=1}^{n}$. 

\texttt{`` Now you need to convert an image description into fine-grained, related visual questions. The
 questions should comprehensively cover detailed visual facts of entities, attributes (e.g., color,
 count, texture, shape, and size), and relationships (e.g., spatial and non-spatial) between the
 entities mentioned in the description. Please complete the task by analyzing each clause in
 the sentence step by step. For each clause, first raise questions about whether each mentioned
 entity exists in the image. Then, raise questions about whether the attributes or relationships
 of the entities are accurately represented in the image. For an image accurately aligned
 with the description, all questions should be answered with “yes”; otherwise, they should be
 answered with “no”.
 Make sure all questions are able to be responded with yes or no and are connected with
 semicolon. Here are examples:
 Example 1:
 description: three black keys, four chickens and a fabric blanket
 output: Are there keys?; Are there three keys?; Are the keys black?; Are there chickens?;
 Are there four chickens?; Is there a blanket?; Is the blanket fabric?
 Example 2:
 description: A person in a blue shirt and red and black apron is using a power tool, likely
 a drill, to assemble a white cabinet or shelving unit indoors. The floor is covered with
 light-colored wood or laminate material.
 output: Is there a person?; Is the person wearing a shirt; Is the shirt blue?; Is the person
 wearing a apron?; Is the apron red and black?; Is the person using a drill?; Is there a white
 cabinet or shelving unit?; Is the person using the drill indoors?; Is there light-colored wood
 on the floor?; Is there laminate material on the floor?
 Example 3:
 description: a large Ferris wheel with a digital clock showing the time as 11:00. The Ferris
 wheel is located in an urban area, as indicated by the modern buildings in the background.
 There is also a tree on the left side of the image, partially obscuring the view of the Ferris
 wheel. The sky appears clear, suggesting a sunny day.
 output: Is there a Ferris wheel?; Is there a digital clock?; Is the digital clock on the Ferris
 wheel?; Is the digital clock showing the time as 11:00?; Is the Ferris wheel located in an
 urban area?; Are there modern buildings in the background?; Is there a tree on the left side?;
 Is the sky clear and sunny?’’}

\paragraph{Generation Prompts.}
These are direct task instructions fed into the diffusion learner $\mathcal{L}_\theta$, specifying the visual or multimodal goal.  
Example templates:
\begin{itemize}
    \item ``Generate a realistic image of a \texttt{\{object\}} in a \texttt{\{scene\}} environment.’’
    \item ``Render a stylized image showing \texttt{\{object\}} with color \texttt{\{attribute\}}.’’
    \item ``Produce a composition of multiple objects: \texttt{\{object\_1\}}, \texttt{\{object\_2\}}, and ensure spatial relation \texttt{\{relation\}}.’’
\end{itemize}
Prompts are sampled from the testing set of GenEval, DPG-Bench, and T2I-bench benchmarks.

\paragraph{Verification Prompts.}
 Verification Prompts is inside the verifier. The input should be the input prompt, generated image and conditions.
 
\texttt{ You are a strict vision-language question validator. 
Return only a compact JSON with fields, please notice that only if average score > 0.95, passed can be true. 
The output format: \textbraceleft passed:boolean, average score:number(0..1), reflection:string, issues:string, must\_fix:string \textbraceright .}
            
Each sub-question will yield a confidence score $s_i^t$ evaluated via cross-modal entailment.  
The structured decomposition generated through condition generation prompt allows the verifier to produce interpretable Yes/No judgments and supports the \textit{No$\rightarrow$Yes} trajectory detection used for DPO data collection.

The generated Json feedback is re-injected into the learner to refine the next iteration.  
All prompts were normalized to maintain stylistic coherence and reduce prompt drift across asynchronous training cycles.


\subsection{Extension to Reasoning Tasks: Math and Coding}
\label{appendix:reasoning}

While SuperIntelliAgent was originally proposed for visual generation, the underlying agentic loop generalizes naturally to symbolic reasoning tasks, including mathematics and code generation.  
Here, the diffusion learner $\mathcal{L}_\theta$ is replaced by a small fine-tunable language model (SLM), and the verifier $\mathcal{V}$ remains a frozen, reasoning-capable LLM.

\paragraph{Math Reasoning.}
For mathematical problems (e.g., AIME or GSM8K), the learner produces an initial chain-of-thought derivation:
\[
\text{``Step 1: … Step 2: … Therefore the result is 18.’’}
\]
The verifier decomposes the prompt into atomic correctness checks such as:
\begin{equation}
\label{eq:swetest2}
\begin{aligned}
c_1 &:~ \text{``Is the algebraic transformation valid?''}, \\
c_2 &:~ \text{``Is the arithmetic consistent?''}, \\
c_3 &:~ \text{``Does the final answer match the problem?''}
\end{aligned}
\end{equation}
Each condition is evaluated by a symbolic parser or by the verifier’s internal reasoning trace.  
Incorrect sub-steps trigger critique feedback—e.g., “You incorrectly expanded $(a+b)^2$.”  
Pairs of incorrect–correct derivations are collected for Direct Preference Optimization, mirroring the No$\rightarrow$Yes progression in image generation.

\paragraph{Coding Tasks.}
For programming tasks (e.g., SWE-Bench), $\mathcal{L}_\theta$ generates candidate patches or function implementations.  
Verifier $\mathcal{V}$ automatically produces structured verdicts barely based on the logical analysis of the coding task itself (no unit tests or API-based simulation needed):
\begin{equation}
\label{eq:swetest}
\begin{aligned}
c_1 &:~ \text{``Does the patch compile?''}, \\
c_2 &:~ \text{``Do all test cases pass?''}, \\
c_2 &:~ \text{``Any corner case that cannot pass?''}, \\
c_3 &:~ \text{``Does the output preserve prior functionality?''}.
\end{aligned}
\end{equation}
Partial failures produce critiques (“Syntax correct but fails edge case 3”) that drive the next refinement iteration.  
This produces verifiable learning signals that allow continual code improvement without human review.

\paragraph{Unified Extension.}
These reasoning tasks reuse the same asynchronous Auto-DPO fine-tuning (Algorithm~\ref{alg:superintelliagent}) without modification.  
Thus, SuperIntelliAgent represents a general-purpose paradigm for self-training across visual, symbolic, and multimodal domains.

Readers may also consult our earlier work, \textbf{CyclePrompt}, for additional inspiration on prompt design \cite{diesendruck2024learning}. In this paper, we mainly concentrate on Image Generation task.

\subsection{Usage in Vicino Product}
\label{appendix:vicino}

The SuperIntelliAgent framework is deployed as the continual learning engine for 2D image generation, within the \textbf{Vicino Creator Suite}, an AI-native 3D and video asset generation platform.  
Here we describe its system-level integration and product impact.

\paragraph{Architecture Integration.}
Vicino uses a modular agent pipeline inspired by Autogen and Semantic Kernel, where each stage (prompt parsing, concept grounding, asset generation, post-verification) corresponds to an agent node.  
SuperIntelliAgent replaces the static generator node.  
During normal user operation, every asset generation request automatically triggers an internal loop:
\begin{enumerate}
    \item The learner synthesizes images textures according to the user’s text input.
    \item The verifier decomposes the request into semantic checks (e.g., geometry correctness, material realism, color fidelity).
    \item If discrepancies are detected, corrective feedback is generated and used to re-render.
    \item Preference pairs are logged asynchronously in the internal replay buffer.
\end{enumerate}
All data remains on-device or within the organization’s private cloud, ensuring privacy-preserving continual improvement.

\paragraph{Continuous Intelligence Growth.}
Every generation cycle contributes to a long-term improvement of Vicino’s visual quality models.  
The diffusion learner continuously absorbs the verifier’s reasoning through lightweight LoRA updates.  
Over time, this leads to emergent specialization—e.g., a studio instance of Vicino gradually adapts to a specific designer’s aesthetic preferences or material library.

\paragraph{Practical Impact.}
This deployment transforms Vicino from a static content generator into an adaptive co-creator that evolves with user behavior.  
Performance evaluation on internal GenEval-style benchmarks shows that iterative Auto-DPO integration improved semantic alignment by $+12.6\%$ and realism preference by $+9.8\%$ over the baseline diffusion model after three days of continual use.  
SuperIntelliAgent thus bridges foundational model training with real-world, production-grade evolution—delivering a form of \emph{productized lifelong learning} in creative AI applications.

\paragraph{Production level Human verifier}
Beyond the automated LLM-based verifier, Vicino incorporates an optional human-in-the-loop verification for production scenarios where aesthetic precision or brand consistency is critical. After each image is generated, users can directly annotate issues—such as geometry inaccuracies, lighting mismatches, or stylistic deviations—and mark their preferred variants. These human preferences are treated as high-quality supervisory signals that complement the LLM verifier: they are logged into the same replay buffer, prioritized during subsequent Auto-DPO optimization, and used to regularize the learner’s updates. 

This hybrid HCI+AI design ensures that Vicino not only self-corrects through automated reasoning but also internalizes user creative judgment, enabling the system to converge toward designs that match real production standards and studio-specific aesthetics.

\subsection{SuperIntelliAgent with Federated Learning}
\label{subsec:federated}
\begin{algorithm}[t]
\caption{Federated SuperIntelliAgent (LoRA-only Aggregation)}
\label{alg:federated}
\begin{algorithmic}[1]
\Require global base model $\theta$, frozen verifier $\mathcal{V}$
\For{each round $r = 1,2,\dots$}
    \For{each client $m$ in parallel}
        \State $\mathcal{D}^{(m)}_{\text{DPO}} \gets$ local Verify--Refine loop
        \State compute LoRA-only update $\Delta\theta^{(m)}$
        \State send $\Delta\theta^{(m)}$ to server
    \EndFor
    \State aggregate $\theta \gets \theta + \frac{1}{M}\sum_{m} \Delta\theta^{(m)}$
    \State broadcast updated LoRA adapters back to clients
\EndFor
\end{algorithmic}
\end{algorithm}
\paragraph{Federated Learning.} While SuperIntelliAgent enables continual self-improvement on a single device, real-world deployment often requires learning across many users and environments without centralizing data.  
To support this setting, we extend the framework into a \emph{federated} architecture, enabling distributed self-supervised preference learning while preserving user privacy.

Let $\{\mathcal{U}_m\}_{m=1}^M$ denote $M$ client devices, each running a local SuperIntelliAgent instance with its own learner $\mathcal{L}_{\theta^{(m)}}$ and verifier $\mathcal{V}$ (shared and frozen).  
Each client autonomously generates preference data using the process in Sec.~\ref{subsec:verification}, producing a local dataset:
\begin{equation}
\mathcal{D}^{(m)}_{\text{DPO}}
= \{ (p_i, x_i^{-}, x_i^{+}) \}_{i=1}^{N_m}.
\label{eq:localdpo}
\end{equation}

Each client performs local DPO fine-tuning:
\begin{equation}
\theta^{(m)} \leftarrow 
\theta - \eta \, 
\nabla_\theta \mathcal{L}_{\text{DPO}}(\theta; \mathcal{D}^{(m)}_{\text{DPO}}),
\label{eq:localupdate}
\end{equation}
and uploads only the LoRA adapter updates $\Delta\theta^{(m)}$, not raw images or preference traces, ensuring privacy-preserving training.  
A central coordinator periodically aggregates:
\begin{equation}
\theta \leftarrow 
\theta + \frac{1}{M} \sum_{m=1}^{M} \Delta\theta^{(m)},
\label{eq:federatedagg}
\end{equation}
similar to FedAvg but applied exclusively on LoRA low-rank adapters to maximize communication efficiency and minimize drift.

\begin{figure*}[t]
    \centering
    \includegraphics[width=\linewidth]{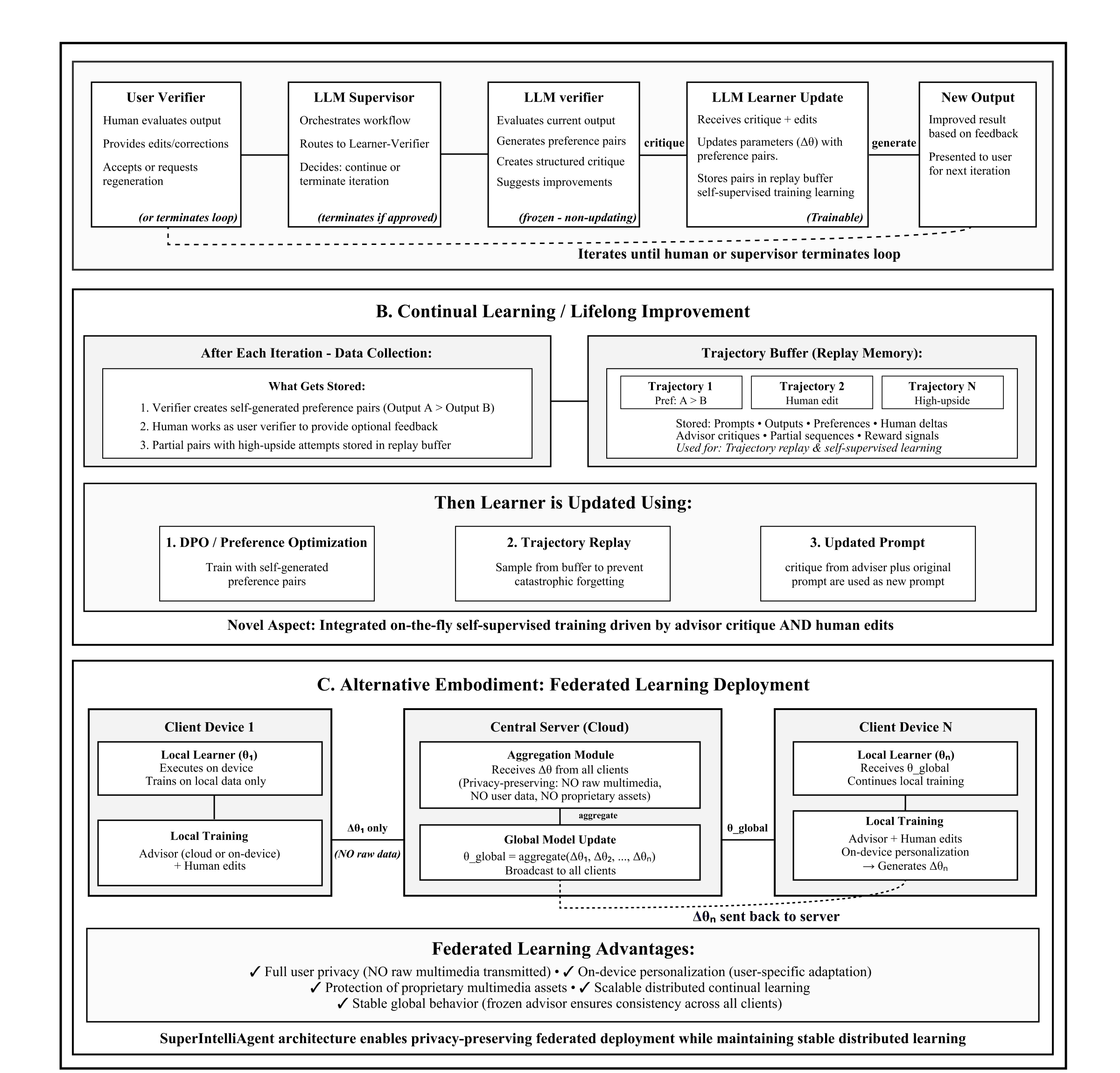}
    \caption{
        Overview of the SuperIntelliAgent pipeline. 
        The learner generates candidate outputs, the verifier performs semantic auditing,
        and DPO-based continual adaptation updates the learner asynchronously.
    }
    \label{fig:flowchart}
\end{figure*}

This federated variant enables SuperIntelliAgent to benefit from the collective diversity of real-world prompts and environments—an essential factor for robust diffusion model alignment—while keeping user data decentralized.  
By sharing only low-rank updates rather than raw examples, the system scales to millions of devices and continuously accumulates global intelligence without compromising privacy or incurring prohibitive communication cost.
The final framework integrating Federated learning and applied in Production in Vicino.ai can be found as in Fig. \ref{fig:flowchart}.

\paragraph{Personalized Adapter.} Federated learning (FL) enables a large model to be collaboratively fine-tuned across many users’ devices, while uploading the adapter weights is not always necessary. Each device can train a small adapter or head on top of the shared base model to capture user-specific data patterns and preferences, all while keeping personal data local for privacy. In an FL setup, a global model would learn general skills from all users, while each user’s adapter fine-tunes the model’s behavior to better suit that user (for instance, adapting a generative model’s image style or text tone to match the user’s preference). Research on federated foundation models indeed explores such dual-personalization approaches: each client learns a global adapter (shared knowledge from the federation) and a local adapter for individual specialization, which are combined at inference time \cite{long2024dual}. This way, the global knowledge ensures a strong base capability, and the personal adapter steers the model to align with the individual’s context or preference (e.g. a personalized slang or image aesthetic). 

\section{LLM-Generated Annotations vs. Human Annotations (Quantity vs. Quality Trade-off)}
SuperIntelliAgent proposes a LLM-Generated Annotation way to generate DPO pairs. We can see that
LLMs can produce a massive number of annotated examples far more quickly than human annotators, dramatically increasing the volume of training data available \cite{baumann2025large}. 

This scalability comes with a caveat: LLM-generated labels often include more noise and false positives. Studies have noted that while LLMs perform well on straightforward annotations, they are prone to mistaken or inconsistent labels in nuanced cases. In fact, one analysis found that as few as 100 carefully curated human annotations (high accuracy, low noise) outperformed a dataset of 1,000 LLM-produced annotations in terms of reducing error rates and spurious findings. A more detailed discussion can also be found in \cite{lin2022rethinking}. The LLM-generated dataset, despite its size, led to significantly higher false-positive rates (10–40\% error) compared to the ~10\% error rate of the small human-labeled set. This “data scale paradox” underscores that more data is not always better if quality is lacking. Choosing between more noisy annotations versus fewer high-precision ones ultimately depends on the application: if false positives or subtle errors must be minimized (e.g. in critical research or safety settings), a smaller but expert-labeled dataset provides more reliable alignment. On the other hand, if speed and coverage are paramount and some noise is tolerable, LLM annotations can bootstrap a model efficiently. An ideal way is to pair LLM annotations with human verification or hybrid strategies to catch critical mistakes to achieve a coarse to fine annotation. In practice, many pipelines now integrate both approaches—leveraging LLMs to efficiently label or rank large datasets, followed by targeted human auditing or refinement to ensure quality. This hybrid strategy is also employed in the Vicino product.

\end{document}